\documentclass{bmvc2k}

\usepackage{svg}
\newcommand{\etal}{\textit{et al}.}
\newcommand{\lpnorm}[1]{\left\lVert #1 \right\rVert}

\newcommand{\Real}{{\rm I\!R}}
\usepackage{tikz}
\usepackage{pgfplots}
\usepackage{multirow, tabularx}
\usepackage{mathtools}

\usepackage{footmisc} 
\usepackage{adjustbox}	
\usepackage{stackengine}

\newcommand\xrowht[2][0]{\addstackgap[.5\dimexpr#2\relax]{\vphantom{#1}}}

\title{Cascaded channel pruning using hierarchical self-distillation}

\addauthor{Roy Miles}{r.miles18@imperial.ac.uk}{1}
\addauthor{Krystian Mikolajczyk}{k.mikolajczyk@imperial.ac.uk}{1}

\addinstitution{
 Imperial College London\\
 Department of Electrical and Electronic Engineering\\
 London, UK
}

\runninghead{Miles and Mikolajczyk}{Cascaded channel pruning}

\def\etal{\emph{et al}\bmvaOneDot}

\begin{document}

\maketitle

\begin{abstract}
In this paper, we propose an approach for filter-level pruning with hierarchical knowledge distillation based on the teacher, teaching-assistant, and student framework. Our method makes use of teaching assistants at intermediate pruning levels that share the same architecture and weights as the target student. We propose to prune each model independently using the gradient information from its corresponding teacher. By considering the relative sizes of each student-teacher pair, this formulation provides a natural trade-off between the capacity gap for knowledge distillation and the bias of the filter saliency updates.
Our results show improvements in the attainable accuracy and model compression across the CIFAR10 and ImageNet classification tasks using the VGG16 and ResNet50 architectures.  We provide an extensive evaluation that demonstrates the benefits of using a varying number of teaching assistant models at different sizes.
\end{abstract}

\section{Introduction}
\label{introduction}

Convolutional neural networks (CNNs) have demonstrated state-of-the-art results on a range of computer vision tasks, such as image classification ~\cite{Simonyan2015VeryRecognition, He2015ResNetRecognition}, depth-estimation ~\cite{Fu2018DeepEstimation, Godard2017UnsupervisedConsistency}, and object detection ~\cite{Girshick2014RichSegmentation, Lin2017FocalDetection}. Despite their success, these models rely on a large number of parameters, which limits their deployment on resource-constrained devices and motivates the need for model compression techniques.
It has been shown that CNNs exhibit significant redundancy, which has led to the development of various pruning techniques. Such methods attempt to identify and remove these redundant weights, which leads to improved memory and computational efficiency with minimal degradation in task accuracy. However, although pruning individual weights~\cite{Lecun1990OptimalDamage, Lee2019SnIP:Sensitivity} can achieve very high levels of sparsity i.e., parameter reduction, the irregular pruning is ill-suited for standard hardware accelerators. In contrast, channel pruning naturally addresses this issue by removing entire convolutional filters.

Most pruning pipelines use rule-based annealing schedules, with intermediate pruning and fine-tuning cycles. We instead jointly train both the set of pruning masks and weights in the same phase. To do this, we first propose a surrogate gradient for the importance scores of each filter, which are updated using standard back-propagation. The binary filter mask is then computed using a global threshold on these scores to achieve a given target compression.
We propose a formulation for the updates of this importance score by extending the idea of "teaching-assistants" (TA) for knowledge distillation~\cite{Mirzadeh2020ImprovedTeacher}. To enable the contribution of previously pruned filters to be re-considered into the student network, we use the gradients from a lesser-pruned TA, thus providing gradients from a model with a higher capacity. We describe the use of passing down surrogate gradients from the TAs to update the pruning masks as \textit{cascaded pruning}, since the pruning is performed in a sequential fashion starting from the largest model in the hierarchy. 

\begin{figure}
    \centering
    \includegraphics[width=.85\linewidth]{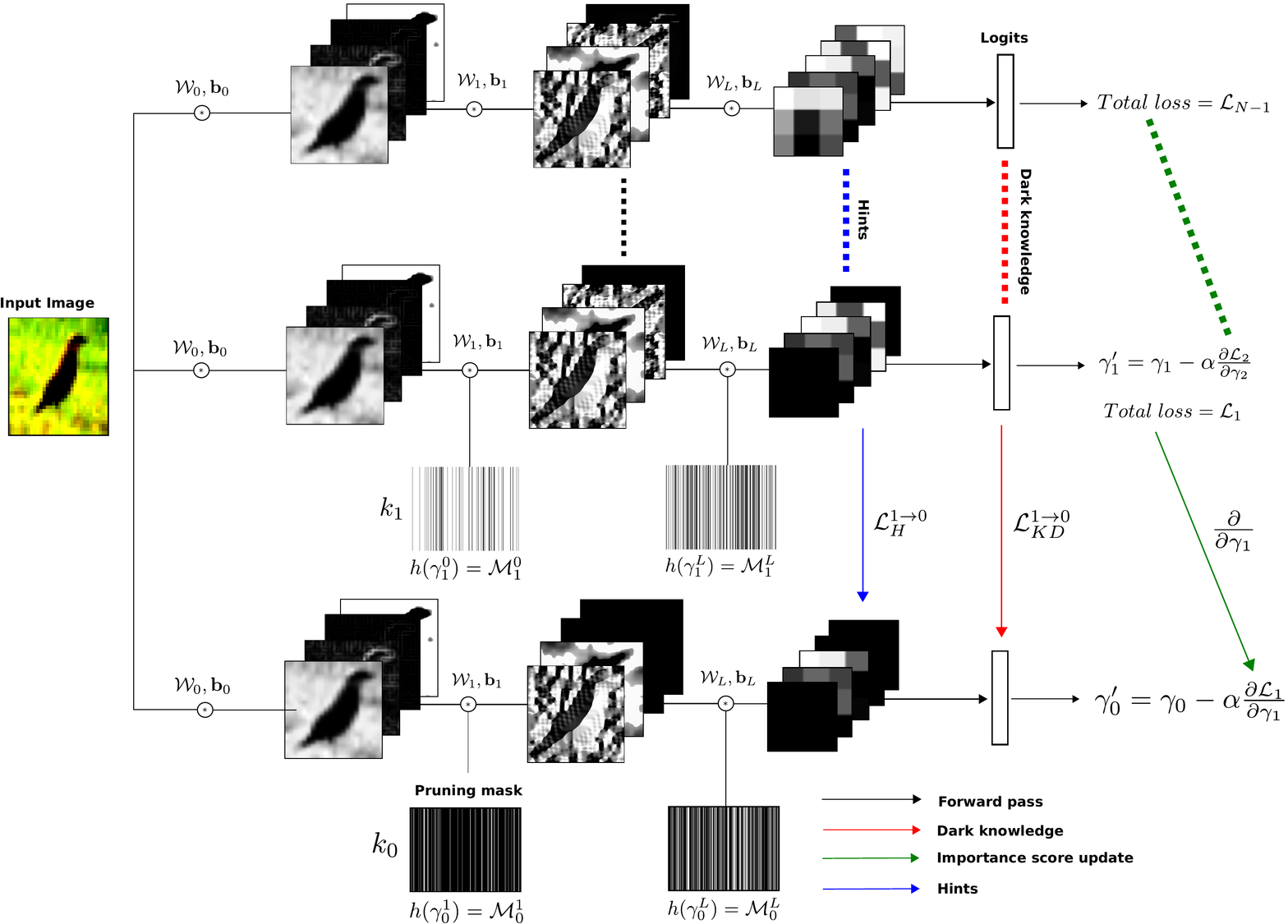}
    \caption{Proposed hierarchical self-distillation strategy for channel pruning. Each of the models are jointly trained with shared convolutional weights but with independent binary masks, batch normalisation layers, and classification layers. The lesser constrained models provide knowledge distillation and importance score gradients down the hierarchy. The frozen teacher for model $T_N$ has been omitted for clarity.}
    \label{fig:multiple_teacher_model}
\end{figure}

Each TA must also share the same set of weights as the student and have the same architecture to reduce the inherent bias in this surrogate gradient term. Using this formulation, we are able to build a hierarchy of student-teacher pairs from the same network (see figure \ref{fig:multiple_teacher_model}) and by considering the relative sizes of each pair, provide a natural trade-off between the bias of the filter saliency gradients and the capacity gap for knowledge distillation. The additional benefit for sharing weights between the student model and all the TA's is a significant reduction in the memory overhead. Having disjoint TA's does not scale well and requires a set of pre-trained models, at appropriate relative sizes, to be readily available. 

We extensively evaluate our approach for widely used state of the art networks and datasets.
For the VGG16~\cite{Simonyan2015VeryRecognition} architecture trained on the CIFAR10 dataset, we are able to achieve a $1.9\times$ reduction in parameters and a $2.3\times$ reduction in FLOPs, while improving upon its top-1 \% accuracy (see table \ref{table:cifar10_results}). We also consider ResNet50~\cite{He2015ResNetRecognition} on the ImageNet2012 classification task, in which we are able to achieve a $3.6\times$ reduction in parameters and a $3.7\times$ reduction in FLOPs for a $2.5\%$ drop in accuracy, which is very significant at this high level of compression (see table \ref{table:imagenet2012_resnet50}).

\section{Related Work}
\label{related_work}

We group the relevant works into three main categories that include filter pruning, knowledge distillation, and efficient architectures.

\noindent{\bf Filter pruning} methods attempt to remove both the feature maps/channels and corresponding filters that have the least positive contribution to the network accuracy. These techniques lead to a structured sparsity in the weights that can directly reduce the number of dense matrix multiplications needed and result in improved on-device performance with standard consumer hardware. 

Pruning filters based on their absolute response magnitude was proposed in \cite{Li2017PruningConvnets}, while \cite{He2017ChannelNetworks} performed the pruning with channel selection based on a LASSO-regression. In~\cite{Luo2017ThiNet:Compression} the pruning of a given layer is guided by subsequent layer statistics. Similarly, NISP~\cite{Yu2018NISP:Propagation} formulates the pruning problem as a binary integer program by which the error-propagation across layers is considered. Discrimination-aware losses were proposed by \cite{Zhuang2018Discrimination-awareNetworks} for selecting channels based on their discriminative power. Probabilistic methods have also been explored for measuring the importance of filters through Bayesian inference and sparsity-inducing priors~\cite{Zhou2018AcceleratePruning, Zhao2019VariationalPruning}. Network pruning can also be modeled as a Neural Architecture Search (NAS) problem, whereby the depth of each layer is incorporated into the design space. AutoSlim~\cite{Yu2019AutoSlim:Numbers} proposed the training of a single slimmable network that is iteratively slimmed and evaluated to ensure minimal accuracy drop, while MorphNet~\cite{Gordon2018MorphNet:Networks} optimizes the model using shrinking and cycle phases. The inefficient filters are then removed using sparsifying regularizers.

In a different approach \cite{Ramanujan2019WhatsNetwork} demonstrated the existence of sparse subnetworks within the large model, with randomly-initialised weights that can achieve high accuracy without any training. They identified this "super mask" using a straight-through estimator for the importance scores of each weight entry. We extend their method in evaluating this mask for the case of pruning entire filters. 





\noindent{\bf Knowledge distillation} 
was originally proposed by ~\cite{Hinton2015DistillingNetwork} to allow a smaller network to learn the correlations between classes from the output of a larger pre-trained teacher model.
This work was extended in ~\cite{Romero2015FitNets:Nets} by using intermediate representations as hints to the student. They approached this by minimising the $L2$ distance between the student and teacher's feature maps. 
%
%
It was shown in \cite{Mirzadeh2020ImprovedTeacher} that the student's performance can degrade if the gap between the student and the teacher is too large. They proposed to use intermediate teaching-assistants to distill knowledge between the teacher and the student. Each of these models used a different architecture and had an independent set of weights. Slimmable neural networks~\cite{Yu2018SlimmableNetworks} defined a network that is executable at different widths through jointly training subsets of uniformly slimmed models. The smaller models benefited from the shared weights and the implicit knowledge distillation provided. We further demonstrate the effectiveness of this knowledge distillation between shared models, but instead independently prune each model using learned pruning masks.


\noindent{\bf Efficient architectures} incorporate the efficiency and accuracy metrics into the initial architectural design choices.
MobileNetV1~\cite{Howard2017MobileNets:Applications} proposed to use depthwise separable layers, which decomposes the 2D convolution operation into two subsequent operations for local spatial and channel aggregation. These layers have been efficiently integrated into all commonly used deep learning frameworks including Inceptions models~\cite{Szegedy2015GoogLeNet/InceptionConvolutions}, and all the MobileNet variants~\cite{Howard2017MobileNets:Applications, Fox2018MobileNetV2:Bottlenecks, Howard2019SearchingMobileNetV3}. 
MobileNetV2~\cite{Fox2018MobileNetV2:Bottlenecks} proposed a linear bottleneck and an inverted residual connection to enforce feature re-use. ShuffleNet~\cite{Zhang2018ShuffleNet:Devices} built upon this idea by using group pointwise convolutions followed by a shuffle operation for enabling cross-group information flow. EfficientNets~\cite{Tan2019EfficientNet:Networks} use compound scaling for uniformly scaling a network's depth, resolution, and width to yield very efficient network architectures. All of these low-rank decomposition methods are complimentary to channel pruning and can be combined with our proposed method for further improvements. We demonstrate this idea in our experiments using the popular MobileNetV1 architecture in section \ref{comparisons_cifar10}.

\section{Method}
\label{method}
In this section, we first provide the formulation for a typical pruning problem and then describe our cascade approach for pruning entire filters through incorporating knowledge distillation, based on the student, teaching-assistant, and teacher paradigm.

\subsection{Formulation}
\label{formulation}
The pruning objective can be described through the use of a binary mask $\mathcal{M} \in \{0, 1\}^{|\mathcal{W}|}$ that is applied to the weights. Although this mask can span all the weights in the network, we restrict our attention to the convolutional layers as they contribute most significantly to the overall computational cost. The objective of pruning is then to learn a small subset of weights that can achieve comparable performance to the original model. These conditions can be described as follows:
\vspace{-.3em}
\begin{align} \label{pruning_conditions}
    \mathcal{L}(f(\mathcal{X}, \mathcal{W} \cdot \mathcal{M}))  \approx \mathcal{L}(f(\mathcal{X}, \mathcal{W})),\;\;\;\;\; 
    \frac{\lpnorm{\mathcal{M}}_0}{|\mathcal{W}|} & = p
\end{align}
Where $p \in [0, 1]$ is a pre-defined pruning ratio that controls the trade-off between the number of used weights, the computational complexity, and the expressiveness of the model.



\subsection{Finding the optimal mask}
\label{finding_optimum_mask}
The binary mask $\mathcal{M}$ disables the least "important" weights. To identify these weights we introduce an importance score $\gamma \in \Real^{|\mathcal{W}|}$. 
This score can be evaluated using a set of static criteria \cite{Lecun1990OptimalDamage, He2018FilterAcceleration} or integrated directly into the learning procedure. The benefit of the latter approach is that the network can capture the complex mutual activations and dependencies of the weights. For example, some of the weights may only be important if another set of weights are enabled or vice-versa.

We model the importance score $\gamma$ as a differentiable weight for which we can compute the binary mask $\mathcal{M}$. A pruning threshold is then defined as the smallest top-$p\%$ of the importance scores across all the convolutional layers. Corresponding weight entries are then conditionally masked if they are below this threshold. 
This operation of mapping the importance scores to the binary mask can be described through the function $h : \Real^{|\mathcal{W}|} \rightarrow \{0, 1\}^{|\mathcal{W}|}$. Since this function is not differentiable with respect to $\gamma$, we adopt the straight through estimator~\cite{Bengio2013EstimatingComputation, Jang2017CategoricalGumbel-Softmax} of its gradients. We also use the derivation from \cite{Ramanujan2019WhatsNetwork} for the $\gamma$ update rule, which we describe below. Consider the following masked convolutional layer:
\vspace{-.3em}
\begin{align} \label{conv_layer}
    \mathcal{Y}_{h, w, n} &= \mathcal{X} * (\mathcal{W} \cdot \mathcal{M}) = \sum_{k_h, k_w}^{K}\sum_{i}^{C} \mathcal{W}_{k_h, k_w, i, n} \cdot \mathcal{X}_{h', w', i} \cdot h(\gamma_{k_h, k_w, i, n})
\end{align}
Where $*$ is the convolution operation and $\cdot$ is the element-wise product. The weights in this equation are represented as a 4-dimensional tensor $\mathcal{W} \in \Real^{K \times K \times C \times N}$, where $K \times K$ is the filter size and $C$, $N$ are the number of input and output channels respectively. The $\gamma$ update can then be computed using the chain rule $\partial \mathcal{L}/\partial \gamma = \partial \mathcal{L}/\partial \mathcal{Y} \cdot \partial \mathcal{Y} / \partial \gamma$.

Due to the previously described practical performance constraint of hardware accelerators, we depart from this general formulation and instead consider pruning entire filters rather than individual weight entries. This results in the binary mask and importance scores for each layer being reduced to $N$-dimensional vectors, where $N$ is the number of filters in the layer. The original ${\partial \mathcal{Y}}/{\partial \gamma}$ term is then also reduced by summing over the spatial and input-channel axes as shown in equation~\ref{dy_dgamma}. In light of this modification, we use $k$ to refer to the ratio of pruned filters, as opposed to $p$ which was used for the ratio of individually pruned weights.
\vspace{-.3em}
\begin{align} \label{dy_dgamma}
    \frac{\partial \mathcal{Y}}{\partial \gamma}_{n} &= \sum_{w}^W\sum_{h}^H\sum_{k_h, k_w}^{K}\sum_{i}^{C} \mathcal{W}_{k_h, k_w, i, n} \cdot \mathcal{X}_{h', w', i} = \sum_{w}^W\sum_{h}^H \mathcal{W} * \mathcal{X}
\end{align}
The final update\footnote{Tensorflow~\cite{Abadi2016TensorFlow:Systems} internally computes all of the gradients and implements these update rules.} for $\gamma$ is given as follows:
\vspace{-.2em}
\begin{align} \label{update_eqn}
    \gamma' = \gamma - \alpha \sum_{w=1}^W\sum_{h=1}^H\frac{\partial \mathcal{L}}{\partial \mathcal{Y}} \cdot (\mathcal{X} * \mathcal{W})
\end{align}
Each filter is assigned an importance score that is related to their weighted contribution in decreasing the task loss $\mathcal{L}$. We expect that incorporating the practical performance metrics into this loss could improve the results, however, we show that using the task loss only is sufficient in providing an excellent accuracy vs model size reduction trade-off.




\subsection{Shared teaching assistants}
\label{shared_teaching_assistants}


An effective use of knowledge distillation requires a set of models with different capacities, strong task accuracy, and with high levels of diversity between them. Each model can then provide supervision to the smaller models through knowledge distillation. In this section, we describe a method for generating this diverse set from the same pre-trained model.

We can uniquely define a model from the same set of weights and architecture through the use of a binary mask $\mathcal{M}$ and its filter pruning ratio $k$. This allows us to define a set of $N$ models $\{T_i \in (\mathcal{M}_i, k_i) \mid 0 \leq i < N\}$ where $k_i < k_{i+1}$ and $k_{N-1} = 1.0$. We expect that the model $T_{i+1}$ is able to distill knowledge to $T_{i}$ since this model has a larger expressive power. We define the $T_0$ model to be the student and $\{T_i \mid 1 \leq i < N-1\}$ to be the set of teaching-assistants (TAs). Model $T_{i+1}$ acts as the teacher for the more constrained model $T_i$, while the teacher for model $T_{N-1}$ is derived from the original pre-trained model with frozen weights.

Each of these models has an independent set of importance scores for each filter, batch normalisation statistics, and classification layers. Independent batch normalisation layers were originally proposed in Yu \etal ~\cite{Yu2018SlimmableNetworks} for networks that are executable at different widths, while the independent classifications layers are needed to enable sufficient diversity between the models. Figure \ref{fig:multiple_teacher_model} shows this proposed cascaded pruning method using a hierarchy of shared TA models.

Sharing the convolutional weights between the teaching-assistants and the student  significantly reduces the training memory overhead while providing implicit knowledge distillation~\cite{Yu2018SlimmableNetworks}. We further conjecture that the set of important filters for model $T_{i+1}$ should also be important for model $T_{i}$ if $k_i \approx k_{i+1}$. Thus, we propose that for each student-teacher pair, we can use the importance score gradients from the teacher to update the student. The proposed modification can be reflected in the $\gamma$ update rule:
%
%
\begin{align} \label{new_update_eqn}
    \gamma'_{i} = \gamma_{i} - \alpha \sum_{w=1}^W\sum_{h=1}^H \frac{\partial \mathcal{L}_{i+1}}{\partial \mathcal{Y}_{i+1}} \cdot (\mathcal{X}_{i+1} * \mathcal{W})
\end{align}
Where the subscript $i$ is used to indicate the $i$th model in the hierarchy. By updating $\gamma_i$ using its corresponding teacher, we are making an assumption that ${\partial \mathcal{L}_{i}}/{\partial \mathcal{\gamma}_{i}} \approx {\partial \mathcal{L}_{i+1}}/{\partial \mathcal{\gamma}_{i+1}}$ or to the very least their sign is the same; which would indicate that the update is moving the importance score in the right direction for $T_i$. The benefit of this proposed formulation is twofold:

\begin{itemize}
  \item Since both the models share the same weights, the teacher's gradient should be a reasonable estimator for the student.
  \item The teacher has a lower pruning ratio $k$ and can thus can provide knowledge distillation to the student.
\end{itemize}


Consider the case where a filter has been pruned away by model $T_{i}$, but not by model $T_{i+1}$. In the original formulation, this will result in the corresponding channels being zeroed out in $\mathcal{X}_{i}$ for the next layer and thus not considered in any of its $\gamma$ updates. In contrast, using the proposed modified update rule from equation \ref{new_update_eqn} will enable the student to consider the weighted contribution of this filter to the loss even if it is currently below the pruning threshold. 





When the TA is much larger than the student, the bias of the gradient estimate will be too large and the updates will span a large set of importance scores, which makes the convergence of a suitable mask for the student very difficult. By ensuring each teacher is sufficiently large to provide useful knowledge distillation, while being sufficiently small such that this gradient update is stable, very effective training can emerge. This result is most noticeable at large pruning rates and evaluated further in section \ref{increasing_number_of_teaching_assistants}.



\subsection{Knowledge distillation}
\label{knowledge_distillation}

Supervised classification uses the cross-entropy loss $H(\cdot, \cdot)$ between the softmax logits of the network $y_s$ and the one-hot encoded ground truth labels $y_{GT}$.
%
%
On the other hand, knowledge distillation uses the KL divergence between the output logits of a teacher model $y_t$ and the student $y_s$~\cite{Hinton2015DistillingNetwork}. The student can then learn the correlations between classes from the teacher's predictions. 
A temperature term $\tau$ can also be used to soften the output probabilities to compensate for the different network capacities.
%
%
The loss for the student is then formulated as the weighted combination of these two terms.
%
%
%
Hinted losses~\cite{Romero2015FitNets:Nets} provide teacher-student supervision for the intermediate representations. For this, we consider the simple reconstruction error term~\cite{Zhuang2018Discrimination-awareNetworks} between feature maps. 



Although sharing of convolutional weights does provide some level of implicit knowledge distillation between models, we find that providing additional explicit knowledge distillation and hints from $T_N$ down to $T_0$ improves the performance of the student. We use the hint losses on the last few layers of the network, while for the knowledge distillation we use the KL divergence between the softened output probabilities of each teacher-student pair. The hyper-parameters $\lambda_{KD}$ and $\lambda_{H}$ are used to scale the KD and hint losses, respectively.


\section{Experiments}
\label{experiments}
In this section, we emperically validate the cascaded pruning approach on CIFAR10, CIFAR100 and ImageNet 2012 classification tasks, in which we consider the VGG16 \cite{Simonyan2015VeryRecognition}. MobileNetV1~\cite{Howard2017MobileNets:Applications} and ResNet50~\cite{He2015ResNetRecognition} architectures respectively. The hint loss is placed in the last 3 layers of VGG16 network and the last 3 residual blocks of ResNet50.
For the first convolutional layer in each network, we do not use any masking and keep an independent set of weights for each model. 

\begin{figure}[ht]
\centering
\subfloat
{
    \adjustbox{width=.41\linewidth}{
    \includegraphics{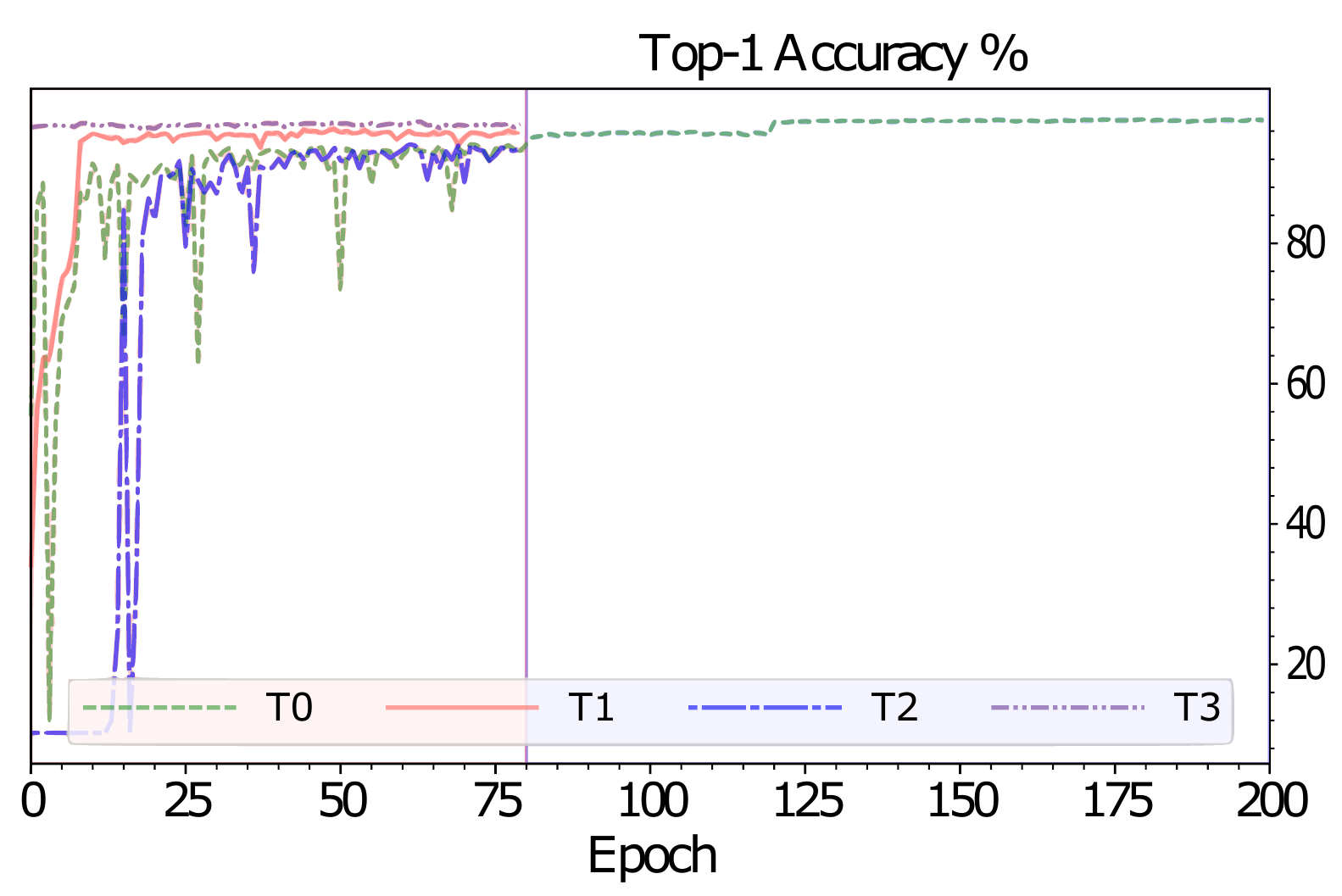}
    }
    \label{fig:training_regimes}
}%
\subfloat
{
    \raisebox{17.5mm}{\adjustbox{width=.57\linewidth}{
\begin{tabular}{|c|c|c|c|c|}
    
    \hline\xrowht[()]{7pt}
    Method & Params & FLOPs  & Top-1 Accuracy (\%) \\
    \hline\xrowht[()]{7pt}
    Original & 3.89M & 495M & 74.10\% \\
    \hline\xrowht[()]{7pt}
    Ours-0.2 & 2.95M & 376M & $\downarrow 0.29\%$ \\
    \hline\xrowht[()]{7pt}
    Ours-0.5 & 1.77M & 156M & $\downarrow 1.42\%$ \\
    \hline\xrowht[()]{7pt}
    Ours-0.7 & 0.38M & 75M & $\downarrow 3.69\%$ \\
    \hline
\end{tabular}%
    }}
    \label{table:mobilenet_cifar100}
}
\caption{Left: Top-1 accuracy's of each model across the joint training step and the fine-tuning step. The training consists of one student $T_0$ and 3 TA's that are trained on the CIFAR10 dataset. Right: Accuracy and performance comparisons on the CIFAR100 dataset using the MobileNetV1 model.} 
\label{fig:training_regimes_and_mobilenet_cifar100}
\end{figure}

The training process is decomposed into two steps; namely, the joint model training and the fine-tuning. The first step consists of jointly training all of the models on the same task, while updating their pruning masks using the proposed formulation in equation \ref{new_update_eqn}. The second step is where we only train the student model with the final frozen binary mask. In this latter stage, the KD and hint loss between the corresponding teacher and the student are still used, but the teacher is replaced with the next subsequent teaching-assistant in the hierarchy  when the student starts to outperforms it on the target task. Figure \ref{fig:training_regimes_and_mobilenet_cifar100} (left) shows these two training steps and we observe that this incremental fine-tuning step is vital to recover the student's accuracy after the previous joint training. The smaller models also experience much higher variance in their validation accuracy during the first stage as the enabled filters are constantly changing. After a suitable number of epochs, these student model naturally converge on a strong mask/weight initialisation for the subsequent fine-tuning step.

For calculating the number of parameters and the computational cost in terms of floating-point operations (FLOPs), we consider just the convolutional layers and the dense layers, without the biases. The batch normalisation layers are not considered as they are typically fused with the previous convolutional weights, while the residual addition and bias terms have a negligible contribution to the total FLOPs. Since we adopt filter pruning, these theoretical FLOP metrics should naturally translate to reduced inference time.




\subsection{Implementation details}
\label{implementation_details}
All our experiments are implemented in Tensorflow \cite{Abadi2016TensorFlow:Systems} with an NVIDIA 2080Ti GPU. We use SGD with Nesterov~\cite{nesterov} as the optimizer with a weight decay of $0.0004$ and momentum of $0.9$. We use a cosine learning rate schedule with an initial learning rate of $0.008$, $5$ epochs per cycle, and an exponential decay. For a given model, we fix its filter pruning ratio $k$ and use $\lambda_{KD} = 0.4$, $\lambda_{H} = 0.001$, and $\tau = 15.0$. Effective knowledge distillation is not only dependant on these loss weights, but also on the relative sizes of each student-teacher pair and the number of intermediate TA's, which is the focus of our attention.

\subsection{Comparisons on CIFAR10 and CIFAR100}
\label{comparisons_cifar10}
The CIFAR10 dataset \cite{Krizhevsky2009LearningImages} consist of 60K $32 \times 32$ RGB images across 10 classes and with a 5:1 training/testing split. The chosen VGG16~\cite{Simonyan2015VeryRecognition} architecture is modified for this dataset by adding independent batch normalisation layers~\cite{Yu2018SlimmableNetworks} after each convolution block and by reducing the number of classification layers to two; of depth $512$ and $10$ respectively. The pre-processing step involves random horizontal flips and center crops of size $32 \times 32$. We jointly train the models for 80 epochs and then fine-tune the student for additional 80 epochs with a batch-size of 128. Each model uses a single student and 3 TAs with the filter pruning ratios of $k_0$, $1 + \frac{k_0 - 1}{1.5}$, $1 + \frac{k_0 - 1}{2.5}$, and $1.0$, respectively. We observe that any reasonably uniform allocation between $k0$ and $1.0$ is sufficient to balance the gradient bias and the capacity gaps.

Table \ref{table:cifar10_results} shows the accuracy and performance metrics for the cascaded pruning  in comparison to other channel pruning methods with the same VGG16 architecture. Our results demonstrate the inherent redundancy in this choice of model for the CIFAR10 task as we are able to achieve significant compression while improving upon its top-1 \% accuracy. 

The CIFAR100 dataset is very similar to CIFAR10, except that it instead contains 100 classes and with 600 images per class. For this evaluation we consider the efficient MobileNetV1 architecture, whereby the learned pruning masks are only applied on the $1\times1$ convolutional layers. We use the same number of TA's and corresponding filter pruning ratios as the CIFAR10 experiments, however, we only fine-tune for the student model for 40 epochs. The results are shown in table \ref{fig:training_regimes_and_mobilenet_cifar100} (right) and demonstrate the validity of this proposed training methodology on an already parameter efficient architecture.

\begin{table*}
    \centering
    \adjustbox{width=.9\linewidth}{
    \begin{tabular}{|c|c|c|c|c|}
        \hline
        Method & Top-1 Baseline Accuracy (\%) & Params & FLOPs  & Top-1 Accuracy (\%) \\
        \hline
        Original & & 14.98M & 313M & 93.26\% \\
        \hline
        Variational pruning \cite{Zhao2019VariationalPruning} & 93.25\% & 3.92M & 190M & $ \downarrow 0.07\%$ \\
        
        Geometric median \cite{He2018FilterAcceleration} & 93.53\% & $-$ & 237M & $ \downarrow 0.34\%$ \\
        
        
        
        Try-and-learn \cite{Huang2018LearningNetworks} & 92.77\% & 2.59M & 140M & $ \downarrow 1.10\%$ \\
        
        Magnitude pruning \cite{Li2017PruningConvnets} & 93.25\% & 5.40M & 206M & $ \downarrow 0.15\%$ \\
        
        

        Discrimination-aware \cite{Zhuang2018Discrimination-awareNetworks} & 93.99\% & 7.80M & 157M & $ \downarrow 
        0.17\%$ \\
        
        
        
        Bayesian Pruning \cite{Zhou2018AcceleratePruning} & 91.60\% & \textbf{0.38M} & 89M & $ \downarrow 0.60\%$ \\
        
        \hline
        Ours-0.3 & 93.25\% & 7.76M & 134M & $\mathbf{\uparrow 0.28\%}$ \\
        Ours-0.6 & 93.25\% & 2.50M & 83M & $\uparrow 0.07\%$ \\
        Ours-0.8 & 93.25\% & 0.97M & \textbf{52M} & $\downarrow 0.28\%$ \\
        \hline
        
    \end{tabular}%
    }
    \smallskip
    \vspace{1em}
    \caption{Comparison to other filter-level pruning methods on the CIFAR10 benchmark and with the VGG16 architecture. Ours-$k0$ indicates a cascaded pruned student model using a filter pruning ratio of k0, whereas the "-" indicates that the results were not reported. For each model, the drop in accuracy is in reference to their own baseline.}
    \label{table:cifar10_results}
\end{table*}

\subsection{Comparisons on ILSVRC-12}
\label{comparisons_imagenet}

We evaluate cascaded pruning on ResNet-50~\cite{He2015ResNetRecognition} for the ImageNet 2012 classification task~\cite{Russakovsky2014ImageNetChallenge}. Unlike most other pruning strategies~\cite{Luo2017ThiNet:Compression, Zhou2018AcceleratePruning}, we also prune the projection layers and the last convolutional layer in each residual block. 
We train the models for 20 epochs and follow this by 20 epochs of student fine-tuning with a batch-size of 30. We use one student and 2 TAs with the filter pruning ratios of $0.3$, $0.5$, and $1.0$, respectively. We follow the same pre-processing step as for the CIFAR10 experiments but with a central crop of size $224 \times 224$. The results can be seen in table \ref{table:imagenet2012_resnet50} and demonstrate strong performance at high-levels of compression.


We observed that using the original SGD optimizer for the $\gamma$ updates led to the majority of the pruning taking place in the last convolutional layer of each residual block, which severely limited the performance improvements. 
This outcome was a consequence of the fixed learning rates across all of the layers and their corresponding importance scores. We replaced SGD with an adaptive learning rate schedule, namely using RMSProp~\cite{rmsprop}, whereby we were able to achieve a more uniform pruning strategy along with faster convergence. To further reduce training time, we also used an additional intermediate fine-tuning step that lasted 10 epochs and consisted of jointly training all the models with fixed pruning masks.

\begin{table}
    \centering
    \adjustbox{width=.9\linewidth}{
    \begin{tabular}{|c|c|c|c|c|c|c|}
        \hline
        & Top-1 Baseline &  &  & Top-1 & Top-5 \\
        Method & Accuracy (\%) & Params & FLOPs & Accuracy (\%) & Accuracy (\%) \\
        \hline
        Original & & 25.5M & 3.86B & 75.03\% & 92.11\% \\ 
        \hline
       
        
        
        
        Discrimination-aware \cite{Zhuang2018Discrimination-awareNetworks} & 76.01 & 12.38M & 1.72B & $\downarrow 1.06\%$ & $\downarrow 0.61\%$ \\
        
        Bayesian Pruning \cite{Zhou2018AcceleratePruning} & 76.10 & - & 1.68B & $\downarrow 3.10\%$ & $\downarrow 2.90\%$ \\
        
        NISP \cite{Yu2018NISP:Propagation} & - & 18.58M & 2.81B & $\downarrow 0.21\%$ & - \\
        
        Filter Sketch \cite{Lin2019FilterPruning} & 76.13 & 14.53M & 2.23B & $\downarrow 1.45\%$ & $\downarrow 0.69\%$ \\
        
        ThiNet \cite{Luo2017ThiNet:Compression} & 72.88 & 12.28M & 1.71B & $\downarrow 1.87\%$ & $\downarrow 1.12\%$ \\
        
        \hline
        & & 25.5M & 4.1B & $\uparrow \mathbf{0.10}\%$ & - \\
        S-ResNet-50~\cite{Yu2018SlimmableNetworks} & 76.10\% & 14.7M & 2.3B & $\downarrow 1.20\%$ & - \\
        \scriptsize{$[0.25, 0.5, 0.75, 1.0]\times$} & & 6.9M & 1.1B & $\downarrow 4.00\%$ & - \\
        & & 2.0M & 278M & $\downarrow 11.10\%$ & - \\
        
        \hline\xrowht[()]{12pt}
        
        
        Cascaded pruning & 75.03 & 7.12M & 1.04B & $\downarrow 2.50 \%$ & $\downarrow 1.29\%$ \\
        
        \hline
    \end{tabular}%
    }
    \smallskip
    \vspace{1em}
    \caption{Top-1 Accuracy and pruning ratios on the ImageNet2012 validation split using the ResNet50 model. The accuracy drops are reported in comparison to their corresponding baseline. The calculations for these baseline performance metrics are covered in the supplementary materials.}
    \label{table:imagenet2012_resnet50}
\end{table}


\section{Ablation studies}
\label{ablation_studies}

In this section we evaluate the benefit of using multiple teaching-assistants. In the supplementary material we further provide a comparison against uniformly pruned baselines and evaluate the impact of using the additional explicit KD loss terms.



\subsection{Increasing the number of teaching-assistants}
\label{increasing_number_of_teaching_assistants}

\begin{figure}[h]
\centering
\subfloat
{
    \includegraphics[width=0.48\linewidth]{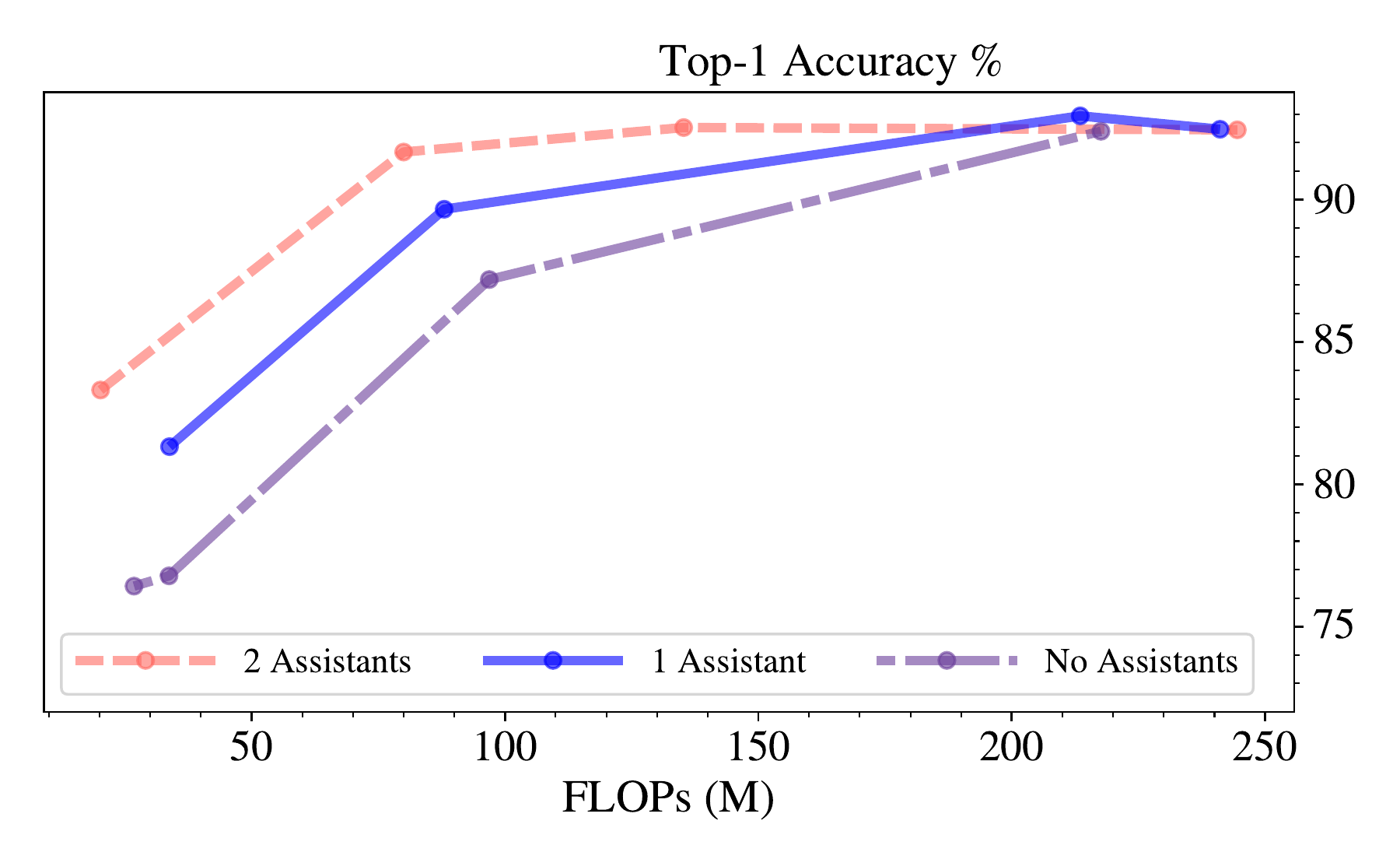}
}%
\subfloat
{
    \includegraphics[width=0.48\linewidth]{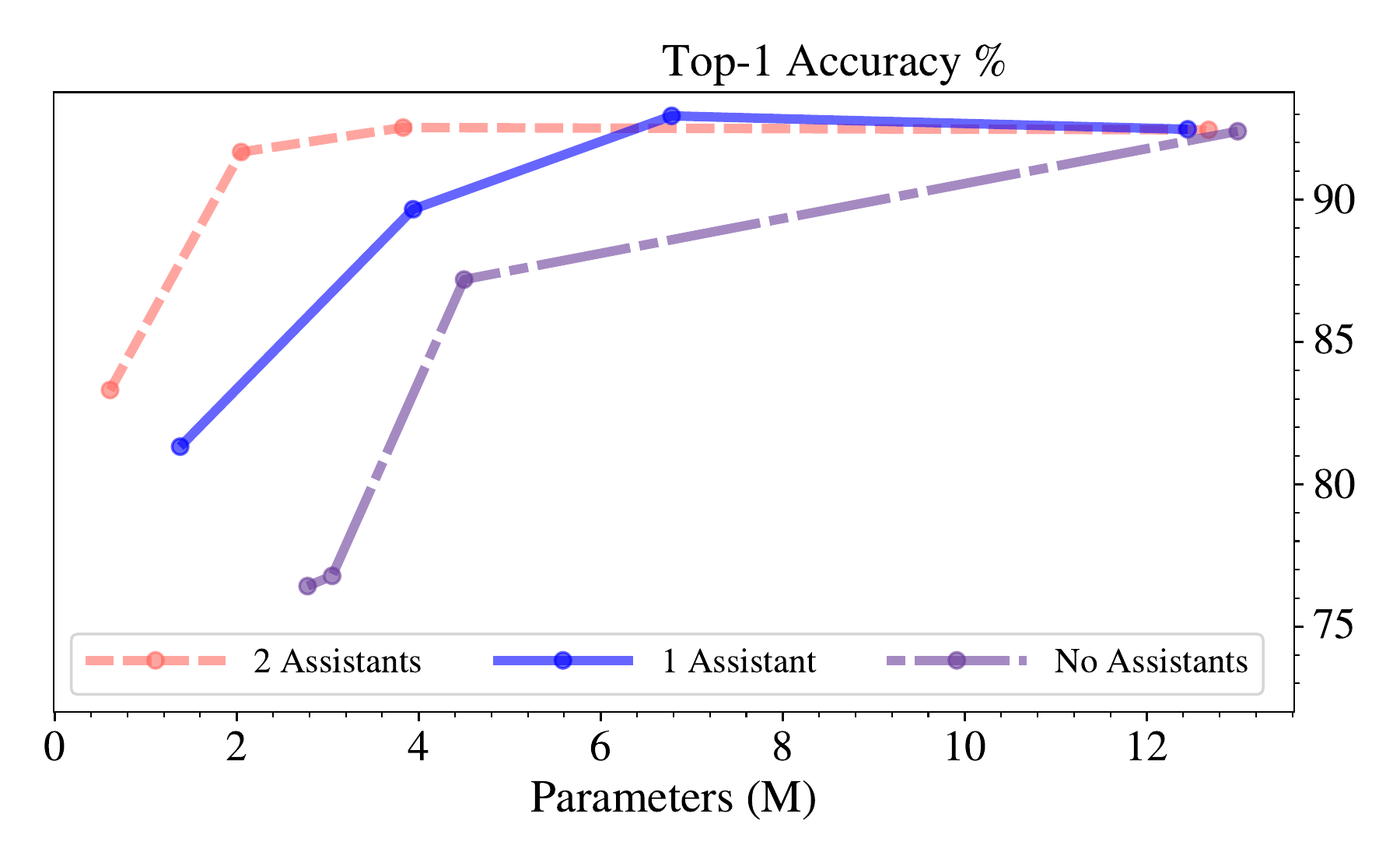}
}
\caption{Evaluation of the computational complexity (left) and the number of parameters (right) for a student model with a varying number of teaching assistants.
Each data point on the graph is ordered according to their pre-defined filter-pruning ratio using the modified VGG16 architecture on the CIFAR10 dataset.} 
\label{fig:ablation_pruning_ratio_and_number_teaching_assistants}
\end{figure}

To evaluate the importance of using the shared teaching-assistant models, we consider training a student with a varying number of teaching-assistants. Figure \ref{fig:ablation_pruning_ratio_and_number_teaching_assistants} shows the task accuracy vs performance trade-offs for training a student with no TAs, with one TA, and with two TAs. In all of these cases, we use the same set of pruning ratios $k$ and when no TA is used, the student only receives knowledge distillation from the fixed pre-trained model. We observe that the student model significantly benefits from having a TA to distill knowledge from and this is especially significant at the higher pruning ratios, whereby the difference in capacities between the student and the teacher is large. 
These results further demonstrate why a uniform allocation of pruning ratios between $k0$ and $1.0$ is most suited for training these models; since it minimises the capacity gap between any of the student-teacher pairs.






\section{Conclusion}
\label{conclusion}

We proposed cascaded-pruning, that is a channel pruning based method using a set of jointly trained and shared models. Each model provides both pruning guidance and knowledge distillation to its corresponding student. Besides the advantages in terms of scalability, cascaded pruning can achieve strong results without any hand-crafted annealing schedules, or iterative training and fine-tuning cycles. We demonstrate these results using a simple straight-through estimator for the pruning mask update, while providing a thorough set of evaluations of their performance with a varying number of teaching-assistants at different sizes. The results are especially significant at high-pruning rates, whereby the student benefits from these intermediate teaching assistants in the fine-tuning stage. We are able to achieve a $\sim 15\times$ compression and $\sim 6 \times$ reduction in FLOPs with negligible accuracy degradation using VGG16 on the CIFAR10 dataset. We also consider the much larger ImageNet dataset with ResNet-50, in which comparable or better accuracy v.s. performance is demonstrated against other state-of-the-art filter pruning methods.
\newline
\newline
\noindent{\bf Acknowledgement.}
This work was supported by UK EPSRC EP/S032398/1 \& EP/N007743/1 grants.
\clearpage



\newpage
\bibliography{references, extra_references}
\end{document}


\maketitle

\section{ResNet50 performance metrics}
Table \ref{table:resnet50_statistics} shows the complexity and parameter break-down for each layer in the ResNet-50 model with an input image of dimensions $224 \times 224 \times 3$.

\begin{table}[ht]
    \centering
    \adjustbox{width=0.6\linewidth}{
    \begin{tabular}{|c|c|c|c|c|c|}
        \hline
        Block \# & $w \times h$ & \#Filters & FLOPs  & Params \\
        \hline
        & $224 \times 224$ & $64$ & 118M & 0.01M \\
        

        \hline
        0 & $56 \times 56$ & $[64, 64, 256]$ & 231M & 0.07M \\
        1 & $56 \times 56$ & $[64, 64, 256]$ & 218M & 0.07M \\
        2 & $56 \times 56$ & $[64, 64, 256]$ & 218M & 0.07M \\
        \hline
        
        3 & $56 \times 56$ & $[128, 128, 512]$ & 295M & 0.38M \\
        4 & $28 \times 28$ & $[128, 128, 512]$ & 218M & 0.28M \\
        5 & $28 \times 28$ & $[128, 128, 512]$ & 218M & 0.28M \\
        6 & $28 \times 28$ & $[128, 128, 512]$ & 218M & 0.28M \\
        \hline
        
        \hline
        7 & $28 \times 28$ & $[256, 256, 1024]$ & 295M & 1.51M \\
        8 & $14 \times 14$ & $[256, 256, 1024]$ & 218M & 1.11M \\
        9 & $14 \times 14$ & $[256, 256, 1024]$ & 218M & 1.11M \\
        10 & $14 \times 14$ & $[256, 256, 1024]$ & 218M & 1.11M \\
        11 & $14 \times 14$ & $[256, 256, 1024]$ & 218M & 1.11M \\
        12 & $14 \times 14$ & $[256, 256, 1024]$ & 218M & 1.11M \\
        \hline
        
        13 & $14 \times 14$ & $[512, 512, 2048]$ & 295M & 6.03M \\
        14 & $7 \times 7$ & $[512, 512, 2048]$ & 218M & 4.46M \\
        15 & $7 \times 7$ & $[512, 512, 2048]$ & 218M & 4.46M \\
        \hline
        
        & $1 \times 1$ & $1000$ & 2.05M & 2.05M \\
        \hline
        \multicolumn{3}{|c|}{Total:} & 3.85B & 25.5M \\
        \hline
        
    \end{tabular}%
    }
    \medbreak
    \caption{Performance statistics for the ResNet50 architecture on the ImageNet2012 dataset.}
    \label{table:resnet50_statistics}
\end{table}

\section{Layer-wise pruning}

\begin{figure}[ht]
\centering
\subfigure
{
    \centering
    \includegraphics[width=.326\linewidth]{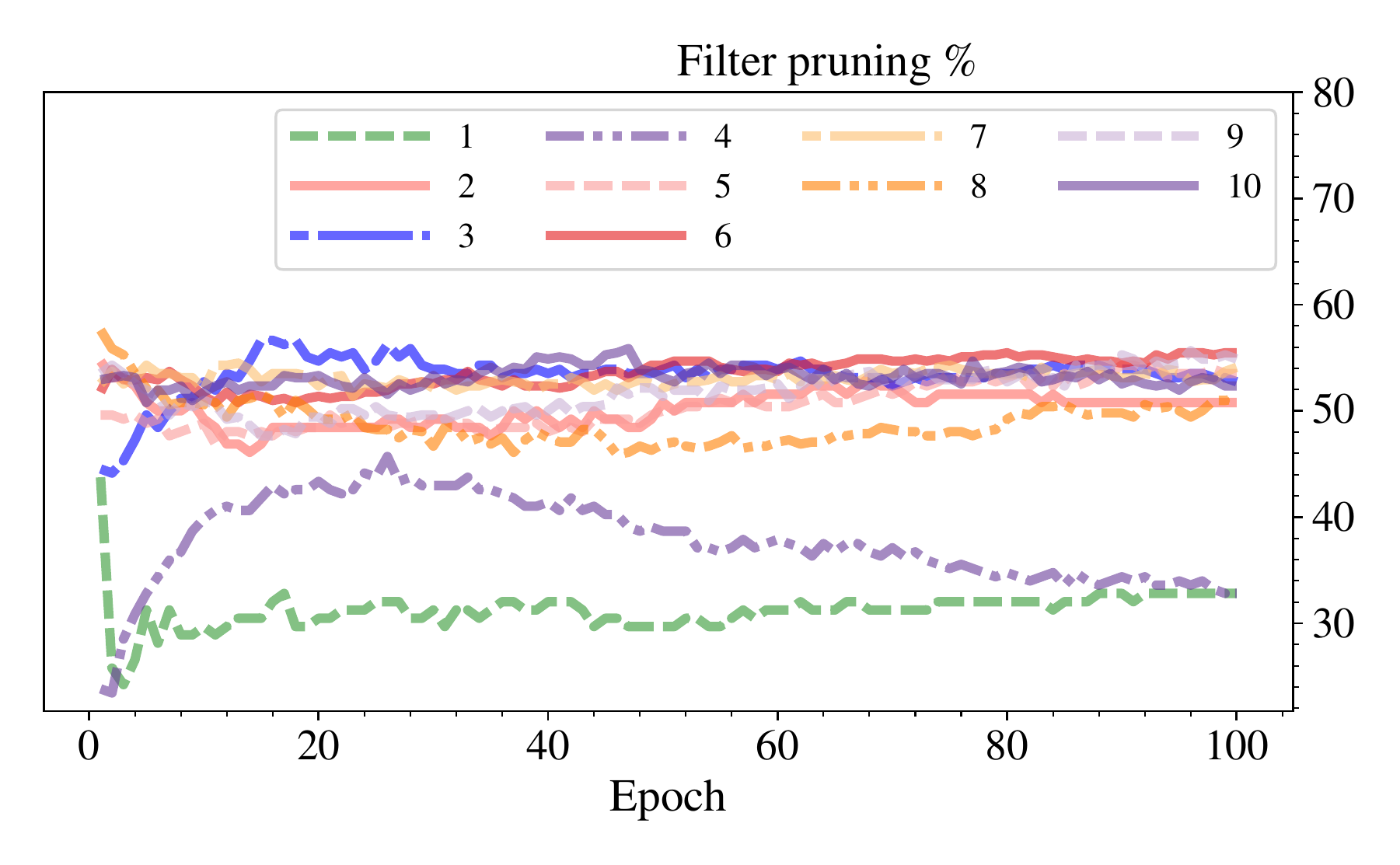}
}%
\subfigure
{
    \centering
    \includegraphics[width=.326\linewidth]{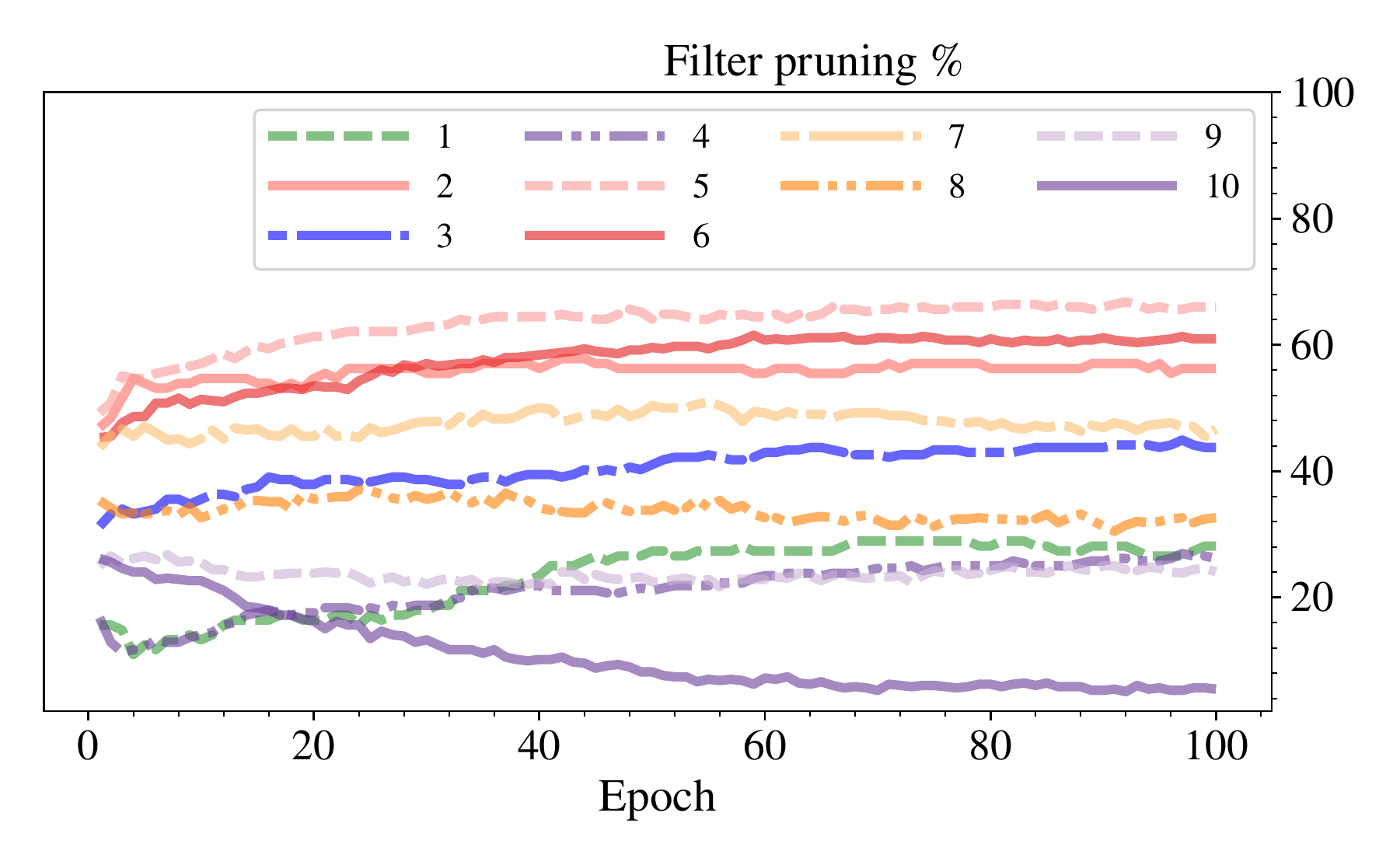}
}%
\subfigure
{
    \centering
    \includegraphics[width=.326\linewidth]{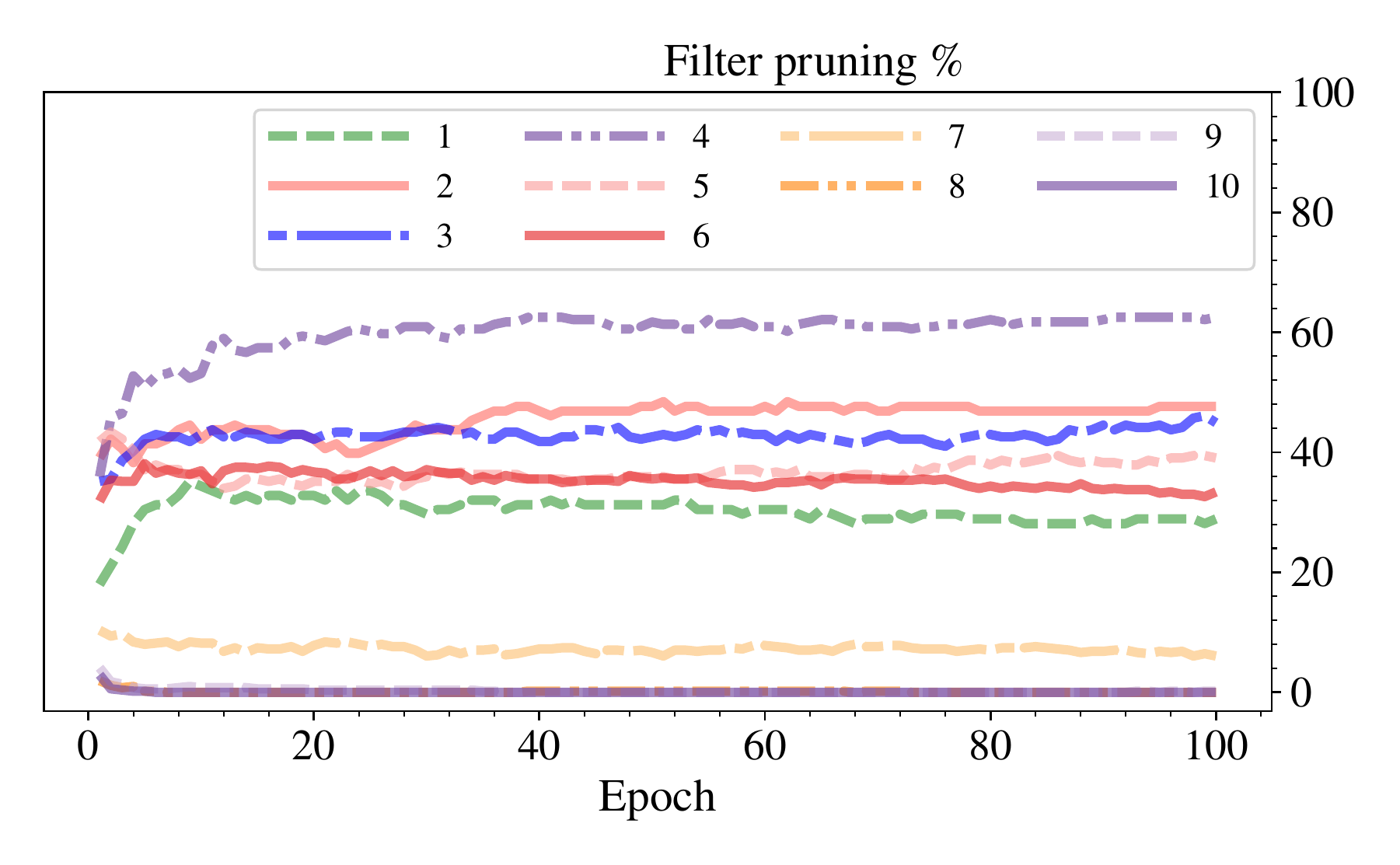}
}%
\caption{The layer-wise pruning for a student and two TA models trained using cascaded pruning. From left to right are models $T_0$, $T_1$, and $T_2$ respectively. Each TA uses the VGG16 architecture and is jointly trained on the CIFAR10 dataset.}
\label{fig:layer_pruning}
\end{figure}

\begin{table}[ht]
    \centering
    \adjustbox{width=.8\textwidth}{
    \begin{tabular}{|c|c|c|c|c|c|c|c|c|}
        \hline
        Layer \# & $w \times h$ & \#Filters & FLOPs  & Params & \multicolumn{4}{c|}{Filter pruning} \\
        \hline
        & & & & & $k_0=0.1$ & $k_0=0.5$ & $k_0=0.6$ & $k_0=0.8$ \\
        \hline
        0 & $32 \times 32$ & 64 & 1.77M & 1.73K    & 0\% & 0\% & 0\% & 0\% \\
        1 & $32 \times 32$ & 64 & 37.75M & 36.86K  & 8.5\% & 20.1\% & 50.8\% & 89.4\% \\
        2 & $16 \times 16$ & 128 & 18.87M & 73.73K & 1.4\% & 11.7\% & 29.6\% & 83.2\% \\
        3 & $16 \times 16$ & 128 & 37.75M & 0.15M  & 40.0\% & 45.4\% & 51.1\% & 67.7\% \\
        4 & $8 \times 8$ & 256 & 18.87M & 0.29M    & 17.8\% & 33.3\% & 45.3\% & 63.1\% \\
        5 & $8 \times 8$ & 256 & 37.75M & 0.59M    & 11.1\% & 36.0\% & 46.1\% & 77.2\% \\
        6 & $8 \times 8$ & 256 & 37.75M & 0.59M    & 37.3\% & 59.0\% & 61.2\% & 86.6\% \\
        7 & $4 \times 4$ & 512 & 18.87M & 1.18M    & 30.1\% & 60.8\% & 61.5\% & 73.7\% \\
        8 & $4 \times 4$ & 512 & 37.75M & 2.36M    & 0\% & 29.7\% & 65.9\% & 67.3\% \\
        9 & $4 \times 4$ & 512 & 37.75M & 2.36M    & 0\% & 17.0\% & 56.5\% & 92.3\% \\
        10 & $2 \times 2$ & 512 & 9.44M & 2.36M    & 6.0\% & 69.0\% & 75.9\% & 88.9\% \\
        11 & $2 \times 2$ & 512 & 9.44M & 2.36M    & 0\% & 85.5\% & 72.4\% & 81.7\% \\
        12 & $2 \times 2$ & 512 & 9.44M & 2.36M    & 0\% & 63.5\% & 63.7\% & 84.1\% \\
        \hline 
        13 & $1 \times 1$ & 512 & 0.26M & 0.26M & 0\% & 31.8\% & 31.9\% & 42.1\% \\
        14 & $1 \times 1$ & 512 & 5.12K & 5.12K & 0\% & 0\% & 0\% & 0\% \\
        \hline 
        
    \end{tabular}%
    }
    \medskip
    \caption{Pruning \% in each layer as a result of cascaded pruning on the CIFAR10 dataset and with the VGG16 architecture at varying filter-pruning ratios. The last two layers (13 \& 14) are the two dense classification layers which are not masked.}
    \label{table:layer_pruning_results}
\end{table}

Figure \ref{fig:layer_pruning} shows the percentage of pruned filters in each layer for a student model and two TAs. The student uniformly prunes the layers, while the larger TA models focus on the last layers. This is in contrast to how most other pruning methodologies work, which tend to result in significant pruning for the last few layers of the network. Table \ref{table:layer_pruning_results} shows how this distribution of pruning levels changes with the filter pruning ratio.

\subsection{Uniformly pruned baselines and KD loss terms}
\label{additional_kd_loss_terms}

The empirical results demonstrated by \cite{Liu2019RethinkingPruning} showed that most channel pruning pipelines achieve comparable or worse performance to training the equivalent smaller model from scratch. Therefore, to confirm the performance benefits of our proposed method, we compare our results against individually training two smaller VGG16 variants from random initialisation. Specifically, we consider using both width scaling and shuffle units~\cite{Zhang2018ShuffleNet:Devices}. Width scaling reduces the depth of each layer by a given \%, while a shuffle unit replaces the convolutional layers with group convolutions and channel shuffles. We use the same training methodology as the original baseline for all these models, which lasts 150 epochs with a cosine learning rate schedule. Liu \etal ~\cite{Liu2019RethinkingPruning} considered two training schemes for these uniformly pruned baseline: training for the same number of epochs as the baseline and training for the same computational budget. In both cases, the reported accuracy's were similar, and in our evaluation we found that further training any of these uniformly pruned baseline results had little effect on the accuracy.

To provide a thorough evaluation of cascaded pruning, we also consider the impact of using the explicit KD and hint loss terms between each student-teacher pair. We use only a single TA with a filter pruning ratio of 0.5 and set $\lambda_{H}=0.001$ and $\lambda_{KD}=0.4$ throughout. These complete sets of results can be seen in table \ref{table:cifar10_scratch_results}.
In the case where no KD or hinted losses are used, only implicit knowledge is distilled between the models, as attributed to the sharing of weights and joint training of all the models. The KD loss term between each teacher-student pair significantly increases the student's performance, while the hinted losses damage the student's performance. The hinted losses perform poorly in this framework since the enabled filters are constantly changing through the importance score updates. However, the student models trained using cascaded pruning still significantly outperform the equivalent smaller models when trained from scratch. These results demonstrate how the learnt mask structure is an integral part and contributing factor to the performance of these cascaded pruned networks. 

\begin{table}
    \centering
    \small
    \adjustbox{width=.7\linewidth}{
    \begin{tabular}{|c|c|c|c|c|}
        \hline
        Method & Params & FLOPs  & Top-1 Accuracy (\%) \\
        \hline
        Baseline & 14.98M & 313M & 93.26\% \\
        \hline
        
        Standard-0.75 & 8.48M & 184M & $\downarrow 5.69\%$ \\
        Standard-0.5 & 3.82M & 89M & $\downarrow 6.82\%$ \\
        Standard-0.25 & \textbf{1.00M} & \textbf{28M} & $\downarrow 11.08\%$ \\
        %
        %
        \hline
        Group-2 & 7.62M & 158M & $\downarrow 6.46\%$ \\
        Group-4 & 3.95M & 80M & $\downarrow 7.69\%$  \\

        \hline
        Ours-0.5 None & 4.13M & 102M & $\downarrow 2.40\%$ \\
        Ours-0.5 w/ KD & 3.62M & 97M & $\mathbf{\downarrow 0.79\%}$\\
        Ours-0.5 w/ Hints & 3.61M & 96M & $\downarrow 2.42\%$ \\
        Ours-0.5 w/ KD \& Hints & 3.69M & 99M & $\downarrow 1.27\%$ \\
        \hline
        
    \end{tabular}%
    }
    \smallskip
    \caption{Accuracy and performance metrics for two efficient VGG16 variants trained from random initialisation on the CIFAR10 dataset. Group-$g$ indicates the use of group convolutions with $g$ groups, while Standard-$s$ uses $s\%$ width scaling for all the convolutional layers.}
    \label{table:cifar10_scratch_results}
\end{table}

\bibliography{references}